\documentclass[final]{tmi99}
\usepackage[dvips]{graphicx}
\usepackage{gb4e}

\title{Explanation-based Learning for Machine Translation}
\author{ Janine Toole\ \ Fred Popowich\ \ Devlan Nicholson\\ \ Davide Turcato\ \ Paul McFetridge}
\institute{Natural Language Laboratory, School of Computing Science,
Simon Fraser University\\8888 University Drive, Burnaby, British
Columbia, V5A 1S6, Canada}

\address{and\\Gavagai Technology\\P.O. 374, 3495 Cambie Street, Vancouver,
British Columbia, V5Z 4R3, Canada}

\email{\{toole, popowich, devlan, turk, mcfet\}@cs.sfu.ca}
\begin{document}
\maketitle

\begin{abstract}
  In this paper we present an application of explanation-based learning (EBL) in the parsing module of a real-time English-Spanish machine translation system designed to translate closed captions. We discuss the efficiency/coverage trade-offs available in EBL and introduce the techniques we use to increase coverage while maintaining a high level of space and time efficiency. Our performance results indicate that this approach is effective.
\end{abstract}

\section{Introduction.}

In this paper we present an application of explanation-based learning (EBL) in the parsing module of a real-time English-Spanish machine translation (MT) system designed to translate closed captions. The core idea of EBL is to convert previous analyses into some 
generalized form that can be called on when similar new examples are encountered (\citeboth{Mitchell:Mit86}; \citeboth{Harmelen:Har88}). The main motivation for using an explanation-based learning approach to parsing is to increase 
efficiency. Dramatic increases in speed can be obtained, at the expense of some coverage and increased memory requirements. In our MT system, efficiency is a major priority because the closed captions included with television and video broadcasts will be translated in real-time. In this paper we discuss the various efficiency/coverage trade-offs that are available within the EBL approach and describe the techniques we use to 
maximize coverage while maintaining efficiency as the primary focus.

This paper is organized as follows. In section 2 we introduce the EBL approach as it has been 
used in natural language processing. Section 3 introduces the translation system in which the 
EBL parser is embedded and discusses the consequences of efficiency versus coverage in this 
domain. In section 4 we introduce our EBL parser and describe the means we use to maximize 
efficiency and coverage. The performance of the system is discussed in section 5. Concluding 
comments can be found in section 6.

\section{\label{tcc}Explanation-based Learning in Natural Language Processing.}

The EBL approach has been extended to natural language processing (NLP) by \citeauthor{Rayner:Ray88} (\citeyear{Rayner:Ray88}), \citeauthor{Srivinas:Sri95} (\citeyear{Srivinas:Sri95}), among others. They use 
existing wide-coverage grammars (or a tree bank \cite{Simaan:Sim97}) to analyze a set of training 
examples. The analyses are then generalized in various ways so that they are applicable to a wide 
range of examples beyond the original input string. In some cases these analyses are manually-checked before generalization to ensure they are the correct interpretation for the given input 
string  \cite{Samuelsson:Sam94}. Typically, the generalized parse is stored with a 
key that is used to subsequently identify when this parse is applicable to a new example. \citeauthor{Srivinas:Sri95} (\citeyear{Srivinas:Sri95}), for example, use the part of speech sequence of the input string as the key. The generalized structures that are saved during EBL training are referred to as `generalized parses,' `macro-rules,' `generalized macro-rules,' or `rule-chunks'. For our purposes these terms are inter-changeable.

It is a characteristic of general wide-coverage 
grammars that they are extremely ambiguous. Hence, parsing involves many options and can be quite slow. The advantage of EBL is that the run-time complexity of parsing is reduced since the majority 
of the parsing occurs off-line. At run-time, the system need only identify the most similar parse 
from those available. In theory the accuracy of the system can be increased since an 
EBL grammar is tuned to the rules actually used in the training domain.

The one disadvantage of the EBL approach is that there is a loss of coverage since the EBL grammar is a subset of the original grammar used to derive it. If an example is not covered by the generalized examples in the EBL grammar then an analysis cannot be provided.

\citeauthor{Rayner:Ray96} (\citeyear{Rayner:Ray96}) note that there are two main parameters that can be adjusted in the EBL learning 
phase: training corpus size and the number and type of macro-rules. They note that the larger the corpus the smaller the loss in 
coverage. However, it should be noted that the size/coverage correlation only holds within a corpus. Comparisons cannot be made across different corpora. For example, \citeauthor{Srivinas:Sri95} (\citeyear{Srivinas:Sri95})
conduct tests on three different corpora. The coverage details of the EBL grammars they developed are given in Table 1. Both the largest and the smallest corpora have extremely poor coverage. 
The median-sized corpus exhibits the best coverage. Hence, the coverage of a given corpus 
depends both on the variability of the structures contained in the corpus as well as on its size.

\begin{table}

\begin{center}

\begin{tabular}{ccc}	\hline		
Corpus & Training Corpus Size(sentences) & Coverage \\ \hline
ATIS & 356 & 80\% \\
IBM  &  1100 & 40\% \\
Alvey  &  80   & 50\% \\ \hline 

\end{tabular}
\caption{Coverage on 3 different corpora, Srivinas \& Joshi 1995}    
\end{center}

\end{table}

The second parameter that \citeauthor{Rayner:Ray96} identify is the number and type of `rule-chunks,' or 
macro-rules, that are generalized. At one end of the spectrum are approaches like \citeauthor{Srivinas:Sri95} (\citeyear{Srivinas:Sri95}) and \citeauthor{Neumann:Neu94} (\citeyear{Neumann:Neu94}) where the whole parse tree for each training example is turned into one 
generalized macro-rule.   This type of grammar is extremely fast, but the coverage loss is typically 
high. At the other end of the spectrum, each rule-chunk is derived from a single rule application. 
This results in a grammar that is identical to the original one; there is no loss in coverage but 
equally no gain in efficiency. \citeauthor{Rayner:Ray94} (\citeyear{Rayner:Ray94}) have taken the approach of trying to find an 
intermediate solution by creating macro-rules corresponding to four possible units: full utterance, recursive NPs, non-recursive NPs and prepositional 
phrases. \citeauthor{Samuelsson:Sam94} (\citeyear{Samuelsson:Sam94}), on the other hand, attempts to identify chunks automatically via an 
entropy minimization method. \citeauthor{Rayner:Ray96} (\citeyear{Rayner:Ray96})  generalize to seven possible units.

The variation in coverage and speed that can be obtained by varying the number and type of rule 
chunks is illustrated by comparing the results of \citeauthor{Srivinas:Sri95} (\citeyear{Srivinas:Sri95}) and \citeauthor{Rayner:Ray96} (\citeyear{Rayner:Ray96}). \citeauthor{Srivinas:Sri95} obtain coverage of 40-80\% (over 3 corpora) with a 60 fold decrease in time when 
compared to parsing their test examples with the original parser and grammar. \citeauthor{Rayner:Ray96} 
claim 95\% coverage with a 10 fold decrease in parsing time.

There is a third parameter, not noted by \citeauthor{Rayner:Ray96} (\citeyear{Rayner:Ray96}), that can be adjusted in EBL. This is the 
amount of information that is retained in the generalized rule. Minimally it is necessary to remove the 
information contributed by the lexical items so that the rule can apply to new word strings. Beyond 
this, there is considerable scope for variation. However, this possibility does not seem to have 
been taken advantage of. \citeauthor{Rayner:Ray94} (\citeyear{Rayner:Ray94}), for example, are typical in keeping all of 
the feature sharing specified in the original rules which make up the generalized macro-rule.

In sum, there are three parameters that can be adjusted in an EBL approach; the size of the training corpus, the size of the rule chunks, and the amount of information that is saved in the generalized parse. Before describing the approach we take, we briefly introduce the MT system in which 
our EBL approach is embedded. This provides the motivation for our efficiency-based approach.

\section{The MT System.}

We have developed a constraint-based lexicalist transfer system, which is designed to 
translate closed captions from English to Spanish \cite{Popowich:Pop97}. 
The intended goal of this 
system is a consumer product that Spanish native speakers will purchase so that they can access North 
American broadcasting in their own language. This requires real-time translation since the translation occurs once the broadcast signal is received by the viewer's television. The translation must 
be produced extremely quickly since the rate of caption flow is quite high. This translation environment dictates that, in the trade-off between efficiency and coverage, efficiency 
should be given the first priority.

There are other factors of the domain which support this preference for efficiency over coverage. 
In the context in which the translations will be used, the captions are just one of several information sources available to the viewer. When reading captions, the viewer also has the sound effects 
and vocal tones of the source language speech, the visual context, and the storyline, each of 
which contribute to convey meaning. Viewers use these information sources to complement the information from the captions. Shortcomings in the translated captions may be made up by the other information sources. Hence, in this domain we can afford to 
sacrifice some coverage in order to achieve the required efficiency. An example of closed captions can be found in Table 2.
\begin{table*}

\begin{center}
\begin{tabular}{l}				\hline
{\scriptsize CYNTHIA, PUSH! COME ON, CYNTHIA!}		\\
{\scriptsize GOOD GIRL.}				\\
{\scriptsize COME ON, CYNTHIA.}				\\
{\scriptsize GOOD GIRL. HE'S GOT THE HEAD.}		\\
{\scriptsize IT'S A BOY.}				\\
{\scriptsize OH! OH!}					\\
{\scriptsize OH, YOU'VE GOT A BOY!}			\\
{\scriptsize HERE, HONEY.}				\\
{\scriptsize HERE YOU GO.}				\\
{\scriptsize OKAY. OKAY.}				\\
{\scriptsize YEAH.}					\\
{\scriptsize SHOULD NOTIFY THE COUNTY, HUH? THANKS.}	\\	\hline
\end{tabular}
\end{center}
\caption{\label{script}Script fragment}
\end{table*}

The translation system consists of analysis, transfer, and generation components as illustrated in Figure 1. Note, this diagram has been simplified somewhat from the details of the actual system. Translation is a non-deterministic multi-phase process: failure in any one process causes back-tracking into a previous process. The rules and lexical entries in the system are defined in terms of feature structures. The analysis and generation phases need not utilise the same theoretical framework. It is only necessary that the appropriate features are used.

The analysis component consists of a part of speech (POS) tagger, a
segmenter, and an EBL parser. The POS tagger uses a
modified subset of the POS tags used in the Oxford Advanced
Learners Dictionary (henceforth OALD). The tagger assigns one tag per
word or phrase. The rule-based segmenter splits an input string into
one or more substrings which are translated separately. We refer to
these substrings as `segments'. 

\begin{figure}[htb]
  \begin{center}
    \includegraphics{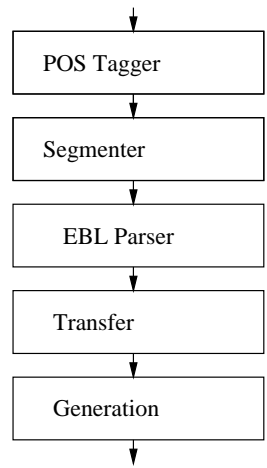}
  \end{center}
  \caption{System architecture}
\end{figure}

\section{An Efficiency-focused EBL Parser.}

In our EBL parser, we use a ``single-rule per training example" approach where the whole parse tree for each training 
segment is converted into one generalized ``macro-rule." Although this maximizes efficiency, it is 
most challenging in terms of coverage since a given generalized example will only apply to examples of the same length (\citeauthor{Srivinas:Sri95} (\citeyear{Srivinas:Sri95}) and \citeauthor{Neumann:Neu94} (\citeyear{Neumann:Neu94}) discuss techniques to 
overcome this limitation). The generalized parses are keyed off the POS tag sequence of 
the input string.

At run-time, the input string is tagged and segmented. Each segment is looked up in the EBL 
index. If the POS sequence is found, then the lexical items from the input string are 
unified with the generalized parse(s) keyed for this POS sequence. If unification is successful, a 
parse has been found and the analysis phase is complete.

While this standard approach meets our efficiency requirements it does
pose some challenges for coverage.  In the following we describe the
techniques we have implemented to minimize coverage loss.

The generalization techniques we describe, whether concerning training
examples, lexical entries or generalized parses, share a common
theoretical motivation, based on the relation between our source
lexicon, our original grammar and our tagger. Our original grammar is
an HPSG-style lexicalist grammar, with most of the syntactic
information encoded in lexical entries. In turn, our lexicon was
derived from the OALD. The core of our lexical entries are macros
corresponding to modified versions of the lexical tags used in the
source lexicon. Each macro defines one relevant lexical class
(e.g. one macro for each verb subcategorization frame) and expands the
information concisely encoded in an OALD tag into a full feature
structure. On the other hand, our tagger uses the same set of
tags. Therefore there is a direct mapping between the lexical macros
used in the lexicon and the tags used by the tagger. This
correspondence has two relevant consequences. The negative consequence
is that our tagset is larger than most standard tagsets (such as that
used in the Penn treebank), and encodes information (like
subcategorization) usually unavailable in other tagsets. This makes
tagging a more challenging task than it is with other tagsets. The
positive consequence of having a lexicalist grammar and a direct
mapping between lexical categories and tags is that our tags encode,
in a nutshell, most of the information relevant to parsing. This
enables us to rely mainly on those tags for parsing and be able to
remove other features from our lexical entries in the generalization
phase without considerably affecting the accuracy of parsing. In other
words, we transfer part of the burden of resolving syntactic ambiguity
on tagging, thus making the parsing task easier.


The first way in which we increase coverage is by performing some of our generalization before 
parsing the training corpus. Before parsing a training example,
we replace each word in the example with the part of speech tag assigned  by our tagger. In addition, we replace the lexicon with one containing a lexical entry for each POS tag. The lexical entry for each of these ``tag-words" is the most general instance of all words 
that can take this tag. For example, the lexical entry for the tag-word ``determiner" is a feature structure corresponding to the generalization of all the individual determiner lexical entries. Having replaced the words with POS tags, we then parse the tag sequence as if it were a regular sentence \cite{Popowich:Pop97}.

This approach allows us to maximize the coverage of our tree. Because the input ``lexical items" 
are extremely general, the parse obtained is quite general and is not a specific idiosyncratic parse 
that may be produced from an input string containing words that are in some way unique or unusual. Hence, 
we end up with parses that are likely to be general and therefore widely applicable to 
new examples.

The second way in which we deal with the coverage issue is by prioritizing the selection of the training 
examples. Most training sets discussed in the literature are quite small (below 5000), the only 
exception being the 15,000 utterance corpus of \citeauthor{Rayner:Ray96} (\citeyear{Rayner:Ray96}). In addition, in many 
cases it is unclear whether the size of the training and test corpora are restricted by the amount 
available or are dictated by some other reason. In any case, if the training corpus is part 
of some larger corpus, there appears to be no principled motivation behind its selection from the 
larger corpus.

In contrast, we prioritize the selection of training examples. In order to maximize the coverage of the grammar, we ensure that our training corpus contains examples that reflect the most common constructions found in the corpus. Our method for extracting the training corpus is as follows, and is illustrated in Figure 2. Our base corpus consists of 11 million words of closed captions. As a first step we segment and tag this corpus to obtain a second corpus which consists of the part of speech sequences for each segment found in the original corpus. We select the 18,000 most frequent POS sequences to comprise our training corpus. Using this approach we guarantee that we are getting the 
most mileage out of our EBL grammar since each training example is attested to be a 
frequently occurring example in this domain. In addition, we increase coverage by using as large a training corpus as possible.

\begin{figure}[htb]
  \begin{center}
    \includegraphics{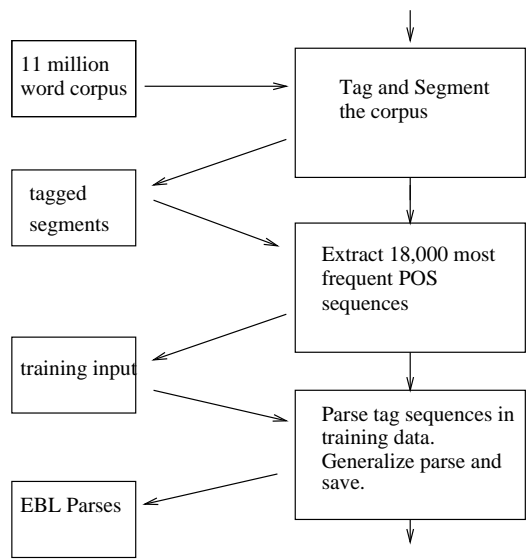}
  \end{center}
  \caption{Training Architecture}
\end{figure}

Once an input sequence has been parsed, we essentially have a generalized parse already, since 
the input string consists of POS tags instead of specific lexical items. In order to maximize coverage, we take our generalization one 
step further and specifically target the parse to reflect the needs of the transfer component of the translation 
system. We ignore the majority of the instantiated features and focus only on the 29 features which are referred to in the MT system's bilingual lexicon. We save any co-indexing found between these features and ignore the specific values of the features for all but eight of the features.

The goal of the bilingual lexicon is to specify mappings between source language lexical items and target language lexical items. Hence, the 29 features which are referenced in the bilingual lexicon, and for which we save co-indexing information, concern the status and indexes of gaps, fillers, modifiers, and complements, features specifying whether the lexical item is interrogative or relative, and features for number, POS, etc. The eight features for which we save values mainly provide details of the status of complements, gaps, fillers, and modifiers in the lexical item, as well as its word type and complement form. Although by saving these features we make our parse less general (the most general form would just save the co-indexing), the realities of our system make this approach the most efficient. These eight features differentiate between many homographic lexical items that have the same part of speech tag, yet have different feature structures. If we do not save the values of these features, then an inappropriate lexical item can unify with the parse. This inappropriate item  is likely to cause problems in either transfer or generation, requiring the system to eventually backtrack into the parser to select the appropriate lexical item. However, to rely on those components to make up for a bad  choice in parsing is time consuming and not the best use of the limited time available to our real-time system. Hence, we save these features in the generalized parse. This means that an inappropriate lexical entry will fail in parsing instead of later in the system. The system can then immediately try other lexical entries which may be more appropriate. A simplified example of a generalized parse can be found in Figure 3.

\begin{figure}[htb]
  \begin{center}
    \includegraphics{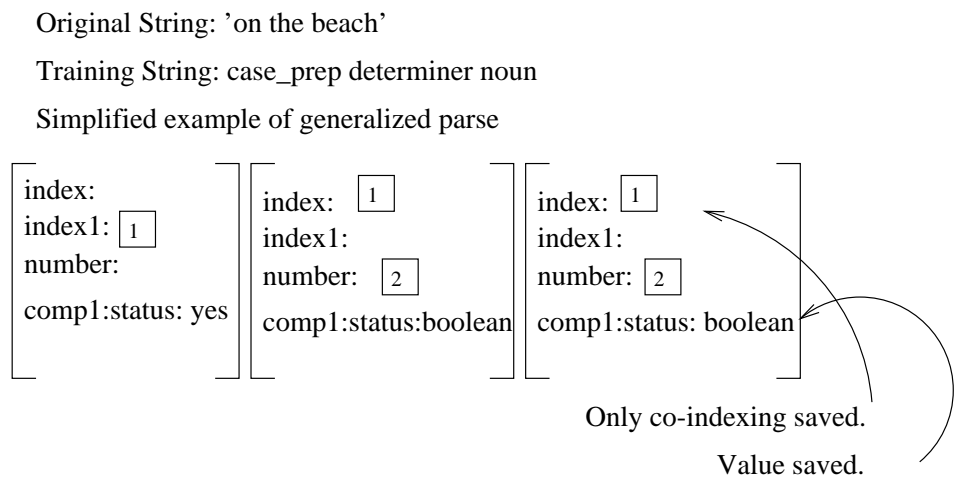}
  \end{center}
  \caption{Simplified Example of a Generalized Parse}
\end{figure}

There are several advantages to this approach. Firstly, our parses are more general since we do not save the specific values of most of the features.  Instead, we save the co-indexing which has been introduced by the  rules that were used in creating the parse. In addition, we are only interested in the co-indexing that holds between the 29 features that are relevant to transfer. For the most part, the values of the features will be instantiated during parsing when the parse is unified with lexical entries. 

Secondly, this approach is more efficient since we are saving less information for each parse. Hence, for an EBL grammar of a specific size, this means an increase in the number of generalized parses that can be saved. Thirdly, fewer features means that run-time unification is faster.


The final technique that we use to increase coverage is to generalize
the key under which parses are saved. For instance, our tagset
categorises prepositions into five classes: noun-modifying,
verb-modifying, verb complement, partitive, and passive
prepositions. Although the features structures for these lexical items
are distinct, in some cases they are combined using the same
rules. Hence, the parses differ only in the values of the features and
not in the co-indexing that parsing introduces. In order to increase
coverage, for selected tags we save parses under a more general tag,
such as preposition. On this approach, a generalized parse for an
input sequence such as `case\_prep determiner noun,' is saved under
the key `preposition determiner noun.' It can then be used to parse
any input sequence containing one of the six prepositions followed by
a determiner and a noun. Hence, the one generalized parse can apply to
a wider range of input. This results in increased
coverage. Furthermore, the resulting parse still contains the more
detailed information from the more specifically tagged
lexical entries. This information is vital for the accuracy of the
transfer module.

In addition to the above approaches, we also include a few run-time heuristics to increase coverage. If a 
generalized parse cannot be found for a POS sequence, we allow the deletion of various lexical 
items. These include adverbs, adjectives, and other parts of speech whose primary goal is to modify information existing in the utterance. After deleting a lexical item, we again search the index 
for the EBL grammar to see if this reduced POS sequence can be found.

It is the characteristics of our domain that allow us to take some of these approaches. The nature of colloquial text is such that the input to the machine translation system may cover a 
very wide semantic domain and may not be strictly grammatical. Hence, our source  grammar is a grammar of colloquial 
English rather than a grammar of formal English. In addition, the grammar is just 
restrictive enough to augment the input lexical entries with sufficient information for the subsequent transfer phase. Any additional information which can be recovered in some other way during transfer is superfluous and thus avoided. For instance, the English grammar does not enforce 
agreement between subject and verb, because the Spanish grammar still has the means (by reference to argument indices) to recognize the subject-verb relation and to independently enforce 
proper agreement on the Spanish output. The unrestrictive nature of our original grammar motivates our unrestrictive approach to our EBL grammar. We maintain only the information that is 
required for effective transfer.

Our heuristic to delete words from the input string is possible because our overall translation goal 
is not ``meaning equivalence" between the input string and its translation, but meaning subsumption 
\cite{Popowich:Pop97}. That is, our translation objectives have been met if the meaning of the target string subsumes the meaning of the source string. This approach is possible because the 
captions are only one of the information sources available to the user.

In sum, our prime objective of efficiency dictates that we use the ``single-rule per training example" approach. We mitigate the loss of coverage engendered by this approach by (i) prioritizing the training corpus, (ii) generalizing the input to training by parsing POS sequences instead of word sequences, and (iii) by reducing the amount of information that is stored for 
each generalized parse.

\section{Performance.}

This section describes a number of experiments carried out to test the effectiveness of our 
approach. We first evaluate the efficiency of the system and then review its coverage.

The EBL grammar  was built from a corpus consisting of the 18,000  most
frequent POS sequences found in the segmented 11 million word
corpus. Around 11,500 of these sequences could be parsed by our regular chart parser and hence the EBL parser covers these 11,500 sequences. Of the sequences which found a parse, the maximum length tag sequence  contains 21 tags and the 
average length is 7.3 tags. The 11,500 sequences which could be parsed were generalized to 8,757 tag sequence keys. Hence, by generalizing the preposition, verb, and auxiliary tags, we are able to reduce our index by about 25\%. This means that in our next phase we can increase the number of tag sequences we use in training.

We limit the number of parses that are stored for each POS sequence in the index. The maximum is a function of the number of generalizable tags in the sequence. Parsing the training corpus resulted in  20,235 generalized parses stored for an average of 2.3 parses per POS index key. The tests  were
carried  out on a file of 1000 sentences randomly selected from the 11
million word corpus.

The  details in Table 3 indicate that the system meets our speed requirements. The average parse time for a segment using 
the EBL approach is significantly faster than our original chart 
parser. The times given in Table 3 reflect the cumulative time spent in the parse module. That is, if the system back-tracks into the parsing module, the additional time spent in this module is added to the total time.

\begin{table}
\begin{center}

\begin{tabular}{lc}	\hline		
Parser & Average time per segment \\ \hline
Chart parser & 900.5 milliseconds \\
EBL Parser  &  4.0 milliseconds \\ \hline 
\end{tabular}
\end{center}
\caption{Parser Comparison}    
\end{table}

Overall, the coverage results are also encouraging. Table 4 provides the details. The EBL database has an
overall coverage of 78.6\% for multi-word segments produced by our segmenter. That is, 78.6\% of the  multi-word segments in the test
file are assigned a
POS tag sequence that it is in the EBL database. Considering that we
take the ``one rule per training example'' approach, this is quite
impressive. 87.9\% of the multi-word segments that were covered by the EBL database
 found a parse in the EBL database (i.e. one of the stored parses unified
successfully with the input lexical items). Of those multi-word segments which
found a parse via the EBL method, 89.9\% subsequently found a translation. In all, 62.1\%  of segments found a translation via the EBL method. This is an increase of 15\% over the results found in our original EBL system. That system provided EBL translations for  54.2\% of the input segments.

\begin{table}
  \begin{center}

  \begin{tabular}{lc} \hline
   Coverage & Performance \\  \hline
    POS sequence found (overall) & \ \ 78.6\%  \\ 
    Parse found (when sequence already found) & \ \ 87.9\% \\ 
    Translation found (when parse already found) &  89.9\% \\ 
    \hline
  \end{tabular}
    
  \end{center}
\caption{Coverage}
\end{table}

In an analysis of 100 of the input sentences, the translations for 59.4\% of the sentences (which consist of one or more segments) were 
rated as acceptable (on an acceptable/ not acceptable scale). In comparison, when translated via 
the original chart parser, an acceptability rate of 68\% was achieved. 
Our criteria for acceptability are discussed in \cite{Popowich:Pop97}.

Some example translations are given in Table 5. Acceptable/Unacceptable translations are identified by 'Y'/'N' respectively. In this table, the notation 'OK' is used for an example which is deemed acceptable but which is not as good as the translation provided by the alternative parser. As expected, there are cases where the coverage of the EBL grammar is less than that of the original grammar. This is illustrated by the examples (a) and (b) which failed to find a parse via the EBL grammar. In the case of (a), the tag sequence corresponding to this input was not found in the EBL index. Hence, no parse could be found. In the case of (b) a parse was found via the EBL approach, but this parse could not be translated by the transfer and generation components of the system. Both examples were translated by the fall-back word-for-word translation method. However, there are also cases where the EBL grammar performs better than the original full grammar. The full grammar failed to find a correct parse for (c) and the input  was translated word-for-word. In contrast, a correct parse was found via the EBL approach, resulting in a more acceptable translation.

\begin{table}
  \begin{center}

  \begin{tabular}{llll} \hline
   &English input &EBL translation&Full Grammar translation\\  \hline
   (a)&why'd you let& por qu\a'e t\a'u alquilando&por qu\a'e lo dejaste ganar-Y\\
   &him win& \a'el victoria-N\\
   (b)&this isn't right &esto es no correcto-OK&esto no est\a'a bien-Y \\ 
   (c)&that's a smart idea& eso es una idea lista-Y&eso es una lista idea-OK\\
    \hline
  \end{tabular}
    
  \end{center}
\caption{Comparison of Output}
\end{table}

These results indicate that the EBL approach will successfully meet
the needs of our real-time English-Spanish translation system. In
addition, they provide clear indication of the areas in which we
should focus to improve performance. For example, the 
coverage  rate of 78.6\% needs to be increased. To do this we plan to integrate our EBL approach with a partial parser.

Secondly, for the cases where a parse is  found, we need to increase
the accuracy of the found parse. We will be analyzing our results to
determine whether the correct parse is not available or if it is
available but not selected. Thirdly, we plan to extend the EBL
approach to the generation phase of the translation system.

\section{Conclusion}

The explanation-based learning approach to parsing provides an
efficient means of providing analyses at the expense of some
coverage. In this paper we described how we took advantage of  these
efficiency gains in order to minimize the analysis time in a real-time
English to Spanish MT  system. We found that even when efficiency is
the prime objective, there are techniques that can  be used to
minimize the coverage loss. On several levels we were
able to maximize  coverage while focussing on efficiency. This was
achieved by making sure that the examples selected for training were
instances of the most frequent constructions, by generalizing the input
so  that the parses produced would be of the most general type, and by
generalizing the saved macro-rules and their key to the minimum needed for
subsequent components of the system.

\bibliographystyle{tmi}

\end{document}